\documentclass{article}

\pdfoutput=1
\newcommand\ms[1]{{\it #1}}
\newcommand\text[1]{{\rm #1}}
\newcommand{\unet}{U-Net}

\usepackage[nonatbib,final]{neurips_2019}

\usepackage[numbers]{natbib}
\usepackage[utf8]{inputenc} % allow utf-8 input
\usepackage[T1]{fontenc}    % use 8-bit T1 fonts
\usepackage{hyperref}       % hyperlinks
\usepackage{url}            % simple URL typesetting
\usepackage{booktabs}       % professional-quality tables
\usepackage{amsfonts}       % blackboard math symbols
\usepackage{nicefrac}       % compact symbols for 1/2, etc.
\usepackage{microtype}      % microtypography
\usepackage{graphicx}       % figures
\usepackage{gensymb}        % \degree
\usepackage{appendix}

\title{Machine Learning for Precipitation Nowcasting from Radar Images}

\author{%
  Shreya Agrawal\\
  \texttt{shreyaa@google.com}\\
  \And
  Luke Barrington\\
  \texttt{lubar@google.com}\\
  \And
  Carla Bromberg\\
  \texttt{cbromberg@google.com}\\
  \And
  John Burge\\
  \texttt{lawnguy@google.com}\\
  \And
  Cenk Gazen\\
  \texttt{bcg@google.com}\\
  \And
  Jason Hickey\\
  \texttt{jyh@google.com}\\
  \AND
  Google Research\\
  1600 Amphitheatre Pkwy\\
  Mountain View, CA 94043\\
}

\begin{document}

\maketitle

\begin{abstract}
High-resolution nowcasting is an essential tool needed for effective adaptation
to climate change, particularly for extreme weather.  As Deep Learning (DL)
techniques have shown dramatic promise in many domains, including the
geosciences, we present an application of DL to the problem of
\emph{precipitation nowcasting}, i.e., high-resolution ($1\text{km} \times
1\text{km}$) short-term (1 hour) predictions of precipitation.  We treat
forecasting as an image-to-image translation problem and leverage the power of
the ubiquitous \unet{} convolutional neural network.  We find this performs
favorably when compared to three commonly used models: optical flow,
persistence and NOAA's numerical one-hour HRRR nowcasting prediction.
\end{abstract}
\section{Introduction}
High-resolution precipitation nowcasting is the problem of forecasting
precipitation in the near-future at high spatial resolutions. This kind of
forecasting requires the processing of large amounts of data at low latency, a
trait well-suited for machine learning.  In contrast, most traditional
approaches use either an \emph{optical flow} (OF) model
\cite{Fleet2006OpticalFE} or a \emph{numerical} model.  OF models attempt to
identify how objects move through a sequence of images, but are unable to
represent the dynamics of storm initiation or decay (which arguably drive most
real-world decisions by those using weather forecasts).  Numerical methods
explicitly simulate the underlying atmospheric physics, and can provide
reliable forecasts, but typically take hours to perform inferences, which
limits their ability to be used in nowcasting.

As weather patterns are altered by climate change, and as the
frequency of extreme weather events increases, it becomes more important to
provide actionable predictions at high spatial and temporal resolutions.  Such
predictions facilitate effective planning, crisis management, and the reduction
of losses to life and property.  A DL-based infrastructure can provide
predictions within minutes of receiving new data, allowing them to be fully
integrated into a highly responsive prediction service that may better suit the
needs of nowcasting than traditional numerical methods.

In this paper, we focus on the subproblem of predicting the instantaneous rate
of precipitation one hour into the future from Doppler radar.  Specifically, we
provide three binary classifications that indicate whether the rate exceeds
thresholds that roughly correspond to \emph{trace rain}, \emph{light rain} and
\emph{moderate rain}. Our forecasts are at $1\text{km}$ spatial resolution, are
within the continental United States and are based on data from
NEXRAD~\cite{nexrad}.  NEXRAD is a network of 159 high-resolution weather radar
stations operated by the National Weather Service (NWS), an agency of the
National Oceanic and Atmospheric Administration (NOAA).

We treat forecasting as an image-to-image translation problem where we are
given a sequence of $n$ input radar images that start at some point of time,
$t_{in_1}$, and end at $t_{in_n}$.  Our task is to generate the radar image
at some point in the future, $t_{out}$.  At the time scales we are working
with, horizontal atmospheric advection is the primary driver for changes in the
radar images, which represent the dynamics we are capturing in our neural
network model.  More specifically, we use the ubiquitous \unet{} Convolutional
Neural Network (CNN)~\cite{unet}.  See the appendices for additional details.
\section{Data setup}

\begin{figure}
  \centering
  \includegraphics[width=0.8\textwidth]{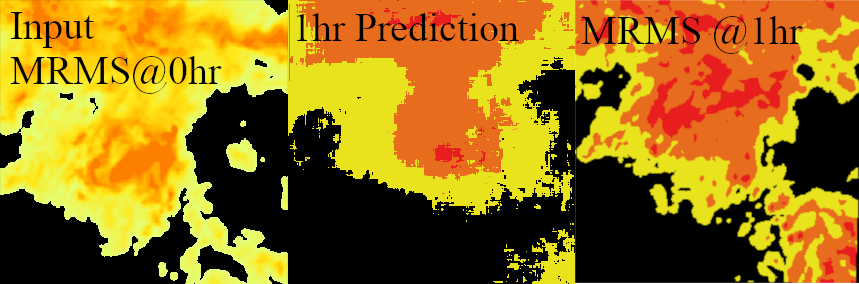}
  \caption{Sample MRMS Image and Predicted Precipitation}
  \label{fig:sample}
\end{figure}

The multi-radar multi-sensor (MRMS) system, developed by NOAA National Severe
Storms Laboratory, provides precipitation rates updated every 2 minutes at a
spatial resolution of $1\text{km} \times 1\text{km}$. The system combines radar
with surface observations and numerical weather prediction methods to get a high
resolution map of current conditions. We use MRMS data for the period of July
2017 through July 2019.

Individually, each radar station scans its environment in a radial pattern
where the scan time and elevation angle is varied to provide a 3D volumetric
reflectivity map. Spatial resolution is generally within 1km radius $\times$ 1
degree azimuth. There are many gaps in coverage but also overlapping regions
covered by multiple stations. We use the MRMS dataset~\cite{mrms}, which
removes non-meteorological artifacts and projects the combined observations
onto a rectangular grid.

We transform the data in three ways. First, for our label images, we quantize
precipitation rates into four discrete ranges based on our three thresholds of
millimeters of rain per hour: $[0, 0.1), [0.1, 1.0), [1.0, 2.5) \text{and}
[2.5, \infty)$. Second, as the US is too large to model at once, we partition
the US into $256\text{km} \times 256\text{km}$ tiles and make predictions for
each tile independently. Third, as most tiles are rainless, we oversample rainy
tiles such that $80\%$ of tiles have at least one pixel of rain.  We trained
our model on data collected in 2018 and tested on the two half-years of data we
had for 2017 and 2019.

Figure~\ref{fig:sample} shows an example of our data. The left image shows the
input. The middle image is our quantized 1-hour nowcasting prediction and the
right image is the quantized one-hour ground truth.

\section{Evaluation and Results}

\begin{figure}
  \centering
  \includegraphics[width=\textwidth]{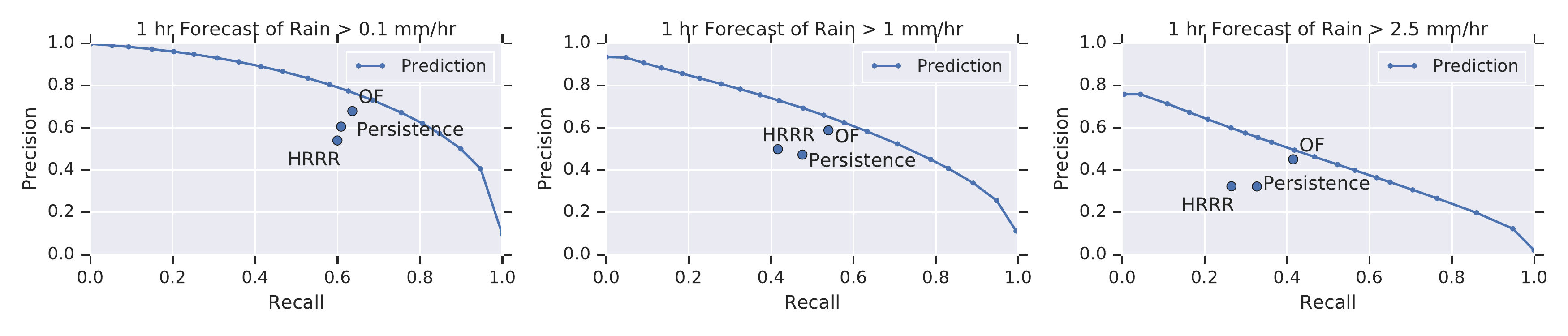}
  \caption{Precision-Recall Curves For Rain Prediction}
  \label{fig:results}
\end{figure}

We evaluate our model on the binary classification performance of our three
different thresholds and treat each output pixel as a separate prediction
when computing precision and recall (PR) metrics. We compare our results with:
\emph{MRMS persistence}, \emph{optical flow} (OF)\cite{gmd-12-4185-2019} and
the \emph{HRRR one hour forecast} \cite{HRRR}.

MRMS persistence is the trivial identity model in which a location is predicted
to be raining in the next hour at the same rate it is raining currently.
Comparing to persistence is common as it can be surprisingly difficult to
outperform.  Optical flow methods are more sophisticated methods that
attempt to explicitly model the velocity of objects moving through a history of
images and are also commonly used in weather forecasting.  HRRR is a
rapid-refresh numerical model from NOAA. It provides 1-hour through 18-hour
forecasts of various atmospheric variables on a 3km grid taking into account
recent satellite and radar observations.  We use a nearest-neighbor heuristic
to align their 3km grid up with our 1km MRMS grid. We use their
\emph{Total\_precipitation\_surface\_1\_Hour\_Accumulation} estimate as a
baseline, which we found to be the best predictor for MRMS among HRRR's various
precipitation forecasts \cite{HRRR}. We only have access to their final
predictions, so we cannot provide full PR curves for their results.

Our model performs better than all three of these methods.  This is
particularly notable when compared to the 1-hour HRRR forecast, which cannot
even be used in practice, as it takes more than an hour to compute.  Instead,
for a 1-hour prediction from \emph{now}, a user would have to use the 3-hour
prediction made 2 hours before \emph{now}, which will yield even worse HRRR
performance than the 1-hour results we are comparing to.  However, once the
prediction window is increased to approximately 5 hours, the HRRR models
consistently outperform our approach.

\section{Future Work}

There are several clear avenues for future work, e.g., the incorporation of
additional modalities of input data such as ground or satellite measurements.
Identifying the most effective means of combining such data in a DL model
remains an active area of research.

Another direction could be refinement on the topological structure and
hyperparameters of the neural network.  In particular, \emph{Generative
  Adversarial Networks} (GANs)~\cite{gan} have shown tremendous promise in
image translation problems where the output is required to have some quality to
make it valid.

Since we perform predictions on independent geographical tiles, border effects
can also be a problem.  When areas of rain exist close to the boundaries of a
tile, the CNN cannot know the direction from which the rain came from, and thus,
where the rain is going to.  Figure~\ref{fig:sample} shows an instance of this
where we do not adequately predict rain in the southeast section of the tile.

There are also many types of additional data that, when combined with radar
data, could significantly extend the utility of our predictions.  E.g., instead
of basing predictions solely on radar data, basing predictions on satellite
data would allow predictions to be made virtually anywhere on the planet.
Indeed, a primary motivation for using CNNs is how simple it is to add and/or
swap out various different images as input.
\section{Conclusion}
We explore the efficacy of treating precipitation nowcasting as an
image-to-image translation problem.  Instead of modeling the complex physics
involved in atmospheric evolution of precipitation, a time consuming and
computational intensive practice, we treat this as a data-driven input/output
problem.  The input is a sequence of MRMS images providing a short history of
rain in a given region and the output is the state of rain one hour afterwards.

We leverage the power of \unet{}s, a type of Convolutional Neural Network
commonly used in image translation problems, and demonstrate that
straight-forward uses can make better predictions than traditional numerical
methods, such as HRRR, for short-term nowcasting predictions presuming the
window for the prediction is on the order of a few hours.  An open question
remains as to whether pure Machine Learning data-driven approaches can
outperform the traditional numerical methods, or perhaps ultimately, the best
predictions will need to come from a combination of both approaches.
\newpage
\appendix
\appendixpage
\section{Related Deep Learning Work}
\label{appendix:DL}
Prior work in applications of DL to precipitation nowcasting falls broadly into
two categories--(1) those that explicitly model time, e.g., with a recurrent
neural network (RNN), and (2) those that use a CNN to transform input images
into a desired output image.

Examples of RNN-based solutions include Shi et al.~\cite{ConvLSTM}, who
introduced the use of convolutional LSTMs (ConvLSTM) which only parameterizes
the more-useful relationships among spatially adjacent regions. Shi et
al.~\cite{Shi17} further improve this by introducing \emph{Trajectory GRU},
which explicitly learns spatially-variant relationships in the data, i.e.,
their model can make different predictions given the same input image based on
features differentiating the geographic location of the input. Sato et
al.~\cite{Sato18} introduce the use of the \emph{PredNet} architecture, which
adds the use of skip-connections and dilated convolutions to further improve
training.

Examples of CNN-based approaches include Lebedev et al.~\cite{yandex}, who also
use a \unet{} architecture. They use their CNN to predict rain at the same
instant as the given satellite image, and then use optical flow algorithms to
make a prediction of future rain. Ayzel et al.~\cite{AYZEL2019186} demonstrate
a baseline CNN model with comparable performance as state-of-the-art optical
flow algorithms. Hernandez et al.~\cite{Hernandez16} also use a CNN to model
the images, but then use a simple perceptron model to perform the nowcasting.
Qui et al.~\cite{Qui17} use a multi-task CNN that explicitly includes features
of the various radar stations to improve their CNN’s quality. We use CNNs not
for estimating rain at the same instant but for nowcasting.

Like Lebedev et al., we experimented with optical flow as well, but unlike
them, we found it performed worse. This is likely because optical flow makes
assumptions that are clearly violated, e.g., the amount of rain will not change
over time.
\section{Problem Formulation}
Ideally, we would estimate a well-calibrated probability distribution of rain
quantities for each pixel: $P(R^{\ms{lat},\ms{lon}}_t \mathrel{|} M_{t-1},
\ldots, M_{t-s})$ where $R^{\ms{lat},\ms{lon}}_t$ is the precipitation rate at
the given latitude and longitude coordinates at time $t$, $M_t$ is the MRMS
data at time $t$, and $s$ is the number of input MRMS images going backwards in
time used as input.  This could be done via Bayesian methods, but such methods
are difficult and often unfeasible in the presence of large quantities of data.

Alternatively, we could perform a regression to come up with the expected
instantaneous rate of rain for each pixel. However, this value has limited
utility as wildly different atmospheric phenomena can yield the same
expectation.  For example, a summer shower might occur with $100\%$ probability
and result in 1mm of rain.  Conversely, a thunderstorm generating 10mm of rain
might be predicted with just a $10\%$ probability.  Both of these events are
\emph{expected} to generate 1mm of rain per hour, but the actions someone would
take in response to these two events are quite different.

So, as a middle ground, we instead provide a series of classifications on
various thresholds of rain: $P_i(R^{\ms{lat},\ms{lon}}_t \ge r_i \mathrel{|}
M_{t-1}, \ldots, M_{t-s})$, where $P_i$ is the probability that the
precipitation rate is at least $r_i$ at time $t$.  This allows us to, e.g.,
explicitly indicate that there is a $100\%$ chance of $1mm/hr$ of rain, and
only a $10\%$ chance of $10mm/hr$ of rain.

\section{Modeling}

Our approach is inspired by the successful application of CNNs to
image-to-image translation. In such tasks, the CNN learns to map an input
image’s pixels to some target image’s pixels. For example, the target image
could explicitly label salient objects in the image, denoise the input, or even
just be the original image itself (in which case the CNN is referred to as an
\emph{autoencoder}). It’s possible to model precipitation nowcasting this way
as well. Given an MRMS image measuring the instantaneous rate of precipitation,
let the target training image be the MRMS image collected one hour after that
instant.

Due to its numerous successes, we use the ubiquitous \unet{}
architecture~\cite{unet}. Like all \unet{}s, ours is a series of
\emph{convolutional blocks} roughly divided into two sections. The first
section, is the \emph{encoder}, and initially applies a basic
\emph{convolutional} block to the image, then iteratively applies several
\emph{downsample convolutional} blocks. The next section is the \emph{decoder},
which takes the output of the encoder, applies a \emph{basic convolutional}
block, followed by a series of \emph{upsampling blocks}. Our three fundamental
convolution blocks are composed of the following operations:

\begin{itemize}
\item Basic Block:
Conv2D $\rightarrow$ BN $\rightarrow$ LeakyReLU $\rightarrow$ Conv2D.

\item Downsample Block:
BN $\rightarrow$ LeakyReLU $\rightarrow$ MaxPooling $\rightarrow$ BN $\rightarrow$ LeakyReLU $\rightarrow$ Conv2D

\item Upsample Block:
Upsample $\rightarrow$ BN $\rightarrow$ LeakyReLU $\rightarrow$ Conv2D $\rightarrow$ BN $\rightarrow$ LeakyReLU $\rightarrow$ Conv2D
\end{itemize}

\emph{Conv2D} stands for a 2D convolution, \emph{BN} stands for Batch
Normalization, and MaxPooling and LeakyReLU are self explanatory. The upsample
operation is resizing via nearest neighbor interpolation.

\emph{Skip-connections} are used to help more efficiently update gradients
during training.  These connections come in two forms. First, \emph{long skip
  connections} are used to connect each \emph{downsample block} in the encoding
phase with a corresponding \emph{upsample block} in the decoding phase. This is
the standard in \unet{}s. Second, \emph{short skip connections} are provided in
every block, as seen in ResNets~\cite{He15} and some \unet{}s as
well~\cite{Drozdzal16}.

We use cross-entropy loss at each pixel in our predictions, and we use ADADELTA
optimization to control our learning rate. We have seven down- and up-sample
blocks; $2 \times 2$ max pooling for downsampling; and 2D convolutions with
kernel size of $3 \times 3$.

We concatenate the MRMS images on the featuremap dimensions where each channel
is a single $256 \times 256$ MRMS tile, collected ten minutes apart over an
hour. For each channel, three additional channels are added: the time of day
the image was taken as well as each pixel’s latitude and longitude. The label
image is the MRMS image collected one hour after the last of the seven input
MRMS images was collected. We used Tensorflow 1.0 as the framework for our
models.
\section{References}
\bibliography{references}{}

\begin{thebibliography}{10}

\bibitem{gan}
Martin Arjovsky, Soumith Chintala, and Léon Bottou.
\newblock Wasserstein gan, 2017.

\bibitem{AYZEL2019186}
G.~Ayzel, M.~Heistermann, A.~Sorokin, O.~Nikitin, and O.~Lukyanova.
\newblock All convolutional neural networks for radar-based precipitation
  nowcasting.
\newblock {\em Procedia Computer Science}, 150:186 -- 192, 2019.
\newblock Proceedings of the 13th International Symposium “Intelligent
  Systems 2018” (INTELS’18), 22-24 October, 2018, St. Petersburg, Russia.

\bibitem{HRRR}
Stanley~G. Benjamin, Stephen~S. Weygandt, John~M. Brown, Ming Hu, Curtis~R.
  Alexander, Tatiana~G. Smirnova, Joseph~B. Olson, Eric~P. James, David~C.
  Dowell, Georg~A. Grell, Haidao Lin, Steven~E. Peckham, Tracy~Lorraine Smith,
  William~R. Moninger, Jaymes~S. Kenyon, and Geoffrey~S. Manikin.
\newblock A north american hourly assimilation and model forecast cycle: The
  rapid refresh.
\newblock {\em Monthly Weather Review}, 144(4):1669--1694, 2016.

\bibitem{nexrad}
Timothy~D. Crum and Ron~L. Alberty.
\newblock The wsr-88d and the wsr-88d operational support facility.
\newblock {\em Bulletin of the American Meteorological Society},
  74(9):1669--1688, 1993.

\bibitem{Drozdzal16}
Michal Drozdzal, Eugene Vorontsov, Gabriel Chartrand, Samuel Kadoury, and Chris
  Pal.
\newblock The importance of skip connections in biomedical image segmentation.
\newblock {\em CoRR}, abs/1608.04117, 2016.

\bibitem{Fleet2006OpticalFE}
David~J. Fleet and Y.~Weiss.
\newblock Optical flow estimation.
\newblock In {\em Handbook of Mathematical Models in Computer Vision}, 2006.

\bibitem{He15}
Kaiming He, Xiangyu Zhang, Shaoqing Ren, and Jian Sun.
\newblock Deep residual learning for image recognition.
\newblock {\em CoRR}, abs/1512.03385, 2015.

\bibitem{Hernandez16}
Emilcy Hernandez, Víctor Sanchez-Anguix, Vicente Julián, J~Palanca, and
  Néstor Duque.
\newblock Rainfall prediction: A deep learning approach.
\newblock In {\em Hybrid Artificial Intelligent Systems}, pages 151--162, 04
  2016.

\bibitem{yandex}
Vadim Lebedev, Vladimir Ivashkin, Irina Rudenko, Alexander Ganshin, Alexander
  Molchanov, Sergey Ovcharenko, Ruslan Grokhovetskiy, Ivan Bushmarinov, and
  Dmitry Solomentsev.
\newblock Precipitation nowcasting with satellite imagery.
\newblock {\em CoRR}, abs/1905.09932, 2019.

\bibitem{gmd-12-4185-2019}
S.~Pulkkinen, D.~Nerini, A.~A. P\'erez~Hortal, C.~Velasco-Forero, A.~Seed,
  U.~Germann, and L.~Foresti.
\newblock Pysteps: an open-source python library  for probabilistic
  precipitation nowcasting (v1.0).
\newblock {\em Geoscientific Model Development}, 12(10):4185--4219, 2019.

\bibitem{Qui17}
M.~{Qiu}, P.~{Zhao}, K.~{Zhang}, J.~{Huang}, X.~{Shi}, X.~{Wang}, and W.~{Chu}.
\newblock A short-term rainfall prediction model using multi-task convolutional
  neural networks.
\newblock In {\em 2017 IEEE International Conference on Data Mining (ICDM)},
  pages 395--404, Nov 2017.

\bibitem{unet}
Olaf Ronneberger, Philipp Fischer, and Thomas Brox.
\newblock U-net: Convolutional networks for biomedical image segmentation.
\newblock {\em CoRR}, abs/1505.04597, 2015.

\bibitem{Sato18}
Ryoma Sato, Hisashi Kashima, and Takehiro Yamamoto.
\newblock Short-term precipitation prediction with skip-connected prednet.
\newblock In {\em ICANN}, 2018.

\bibitem{ConvLSTM}
Xingjian Shi, Zhourong Chen, Hao Wang, Dit{-}Yan Yeung, Wai{-}Kin Wong, and
  Wang{-}chun Woo.
\newblock Convolutional {LSTM} network: {A} machine learning approach for
  precipitation nowcasting.
\newblock {\em CoRR}, abs/1506.04214, 2015.

\bibitem{Shi17}
Xingjian Shi, Zhihan Gao, Leonard Lausen, Hao Wang, Dit-Yan Yeung, Wai-kin
  Wong, and Wang-chun WOO.
\newblock Deep learning for precipitation nowcasting: A benchmark and a new
  model.
\newblock In I.~Guyon, U.~V. Luxburg, S.~Bengio, H.~Wallach, R.~Fergus,
  S.~Vishwanathan, and R.~Garnett, editors, {\em Advances in Neural Information
  Processing Systems 30}, pages 5617--5627. Curran Associates, Inc., 2017.

\bibitem{mrms}
Jian Zhang, Kenneth Howard, Carrie Langston, Brian Kaney, Youcun Qi, Lin Tang,
  Heather Grams, Yadong Wang, Stephen Cocks, Steven Martinaitis, Ami Arthur,
  Karen Cooper, Jeff Brogden, and David Kitzmiller.
\newblock Multi-radar multi-sensor (mrms) quantitative precipitation
  estimation: Initial operating capabilities.
\newblock {\em Bulletin of the American Meteorological Society},
  97(4):621--638, 2016.

\end{thebibliography}
\bibliographystyle{plain}
\end{document}